\documentclass[10pt,conference]{IEEEtran}
\usepackage{amsmath,amsfonts,mathtools}
\usepackage{algorithmic}
\usepackage{algorithm}
\usepackage{array}
\usepackage{balance}
\usepackage{textcomp}
\usepackage{stfloats}
\usepackage{url}
\usepackage{verbatim}
\usepackage{graphicx}
\usepackage{cite}
\usepackage{enumitem}
\usepackage{amsfonts}
\usepackage{caption}
\usepackage{subcaption}
\usepackage{todonotes}
\usepackage[outdir=./]{epstopdf}
\usepackage{hyperref}
\hypersetup{%
  colorlinks=true, 
  breaklinks=true, 
  urlcolor=blue, 
  linkcolor=blue, 
  citecolor=blue, 
} 

\usepackage{amsthm}
\newtheoremstyle{mystyle}
  {}
  {}
  {\itshape}
  {}
  {\bfseries}
  {.}
  { }
  {}

\theoremstyle{mystyle}

\newtheorem{problem}{Problem}

\begin{document}
\bstctlcite{IEEEexample:BSTcontrol}
\title{V2N Service Scaling with\\Deep Reinforcement Learning}

\author{

\IEEEauthorblockN{Cyril Shih-Huan Hsu\IEEEauthorrefmark{1},
Jorge Martín-Pérez\IEEEauthorrefmark{2},
Chrysa Papagianni\IEEEauthorrefmark{1} and
Paola Grosso\IEEEauthorrefmark{1}  
}


\IEEEauthorblockA{
\IEEEauthorrefmark{1}\textit{Informatics Institute, University of Amsterdam}, The Netherlands \\
\IEEEauthorrefmark{2}\textit{Departamento de Ingeniería Telemática, Universidad Carlos III de Madrid}, Spain
\\ s.h.hsu@uva.nl, jmartinp@it.uc3m.es, c.papagianni@uva.nl, p.grosso@uva.nl
}


}




\maketitle

\begin{abstract}
The fifth generation (5G) of wireless networks is set out to meet the stringent requirements of vehicular use cases. Edge computing resources can aid in this direction by moving processing closer to end-users, reducing latency. However, given the stochastic nature of traffic loads and availability of physical resources, appropriate auto-scaling mechanisms need to be employed to support cost-efficient and performant services. To this end, we employ Deep Reinforcement Learning (DRL) for vertical scaling in Edge computing to support vehicular-to-network  communications.
We address the problem using Deep Deterministic Policy Gradient (DDPG). As DDPG is a model-free off-policy algorithm for learning continuous actions, we introduce a  discretization approach to support discrete scaling actions. Thus we address scalability problems inherent to high-dimensional discrete action spaces. 
Employing a real-world vehicular trace data set, we show that DDPG outperforms existing solutions, reducing (at minimum) the average number of active CPUs by 23\% while increasing the long-term reward by 24\%.

\end{abstract}

\begin{IEEEkeywords} V2N, scaling, DRL, DDPG, A2C
\end{IEEEkeywords}

\section{Introduction}
\label{intro}


Connected and Automated Vehicles (CAVs) is a transformative technology for the automobile industry. 
CAV applications (real-time situational awareness etc.) require process-intensive and low-latency, reliable computing and communication services. Such characteristics prohibit the use of cloud computing resources that are usually centralized into large data centers. An effective approach to address latency requirements is to leverage Edge computing,  moving computing resources closer to where the data is being generated, processed, and consumed.

Due to the ubiquity of the cellular infrastructure, 5G systems are set out to support Cellular Vehicle-to-Everything (C-V2X) communications, ensuring ultra-low latency and ultra-high reliability communications (URLLC) under high-density and -mobility conditions. The C-V2X technology, introduced by 3GPP~\cite{3GPPV2X1}, refers to the low-latency communication system between vehicles and vehicles (V2V), pedestrians (V2P), roadside infrastructure (V2I), and cloud/edge servers (network, V2N). 
Each of these use cases has different communication requirements. 
5G systems are expected to address such requirements by slicing the physical network into several tailor-made logical ones e.g., for autonomous-driving, tele-operated driving etc \cite{wijethilaka2021survey}. 
In this ecosystem, Edge computing is employed to support dynamic service creation and processing per slice. 

Nevertheless, appropriate mechanisms must be put in place to ensure elastic network services in order to meet service level agreements.
Broadly speaking, elasticity is the ability to increase and shrink selected resources in a systematic and autonomous manner to adapt to workload changes \cite{medeiros2020enabling}.
Similar to  \cite{de20215growth}, using the vehicular traffic from the streets of Turin, we dynamically scale vertically Edge computing resources, to accommodate the latency requirements for V2N applications.
However, in this work we employ DRL for deciding on how to vertically scale computing resources. 

The strength of RL approaches lies in their ability to reason under uncertainty and adapt to changes at runtime, which maps well onto the stochastic V2N environment. RL has been investigated before for scaling computing resources;
 authors in \cite{verma2021auto, autopilot, k8s, derp} provide ML/RL-based auto-scaling techniques in the context of cloud resource management. 
ML has been also employed for scaling virtualized network functions \cite{subramanya2019machine,rahman2018auto, gari2021reinforcement, 203302,chai2018trafficaware,7417181}. For instance, DDPG has been utilized to predict a threshold
vector of CPU loads that eventually triggers scaling, but not the scaling actions per se \cite{chai2018trafficaware}. Authors in~\cite{7417181} formulate the scaling of computing resources as a Markov
Decision Process (MDP) 
 and propose an RL approach based on Q-Learning. 
However to enable flexible scaling decisions that can support surges in network traffic, we need to go beyond approaches that employ a limited action space, as such solutions often scale CPU in increments of one. In consequence, this would lead to scalability problems due to the high-dimensional discrete action space. 
Instead, we propose the use of a DRL approach with continuous action space, introducing a discretization method supporting the scaling actions.
Our contributions of are summarized as follows:
\setlist{nolistsep}
\begin{itemize}[noitemsep,leftmargin=*]
    \item We investigate the use of DDPG for the V2N scaling problem, 
    introducing a discretization method termed as \textit{Deterministic Ordered Discretization} (DOD), forming the \textit{DDPG-DOD} approach. The DOD method
    can be further applied to off-the-shelf RL algorithms with continuous action space to address discrete problems with ordering properties.
    \item We compare the performance of the proposed DRL agents with the traditional, prediction and RL algorithms presented in \cite{de20215growth}, using road traffic traces from Turin. Furthermore we compare against  Advantage Actor Critic (A2C) \cite{10.5555/3045390.3045594}, a discrete DRL approach for scaling resources \cite{beks21}. A2C  has been selected as it outperforms respective DRL methods (i.e., Deep Q-Network) 
    in a variety of RL benchmarks\cite{10.5555/3045390.3045594}. 
\end{itemize}

\noindent In the following, we  describe the V2N system in
{\S\ref{sec:system}}.
Then, we model the scaling problem as an MDP 
and introduce 
DDPG-DOD to address it
in {\S\ref{sec:problem}}.
Finally, we evaluate the
proposed approach ({\S\ref{sec:results}})
and highlight our conclusions ({\S\ref{sec:conclusions}}).

\section{V2N system description}
\label{sec:system}

\begin{figure}[t]
    \centering
    \includegraphics[width=\columnwidth]{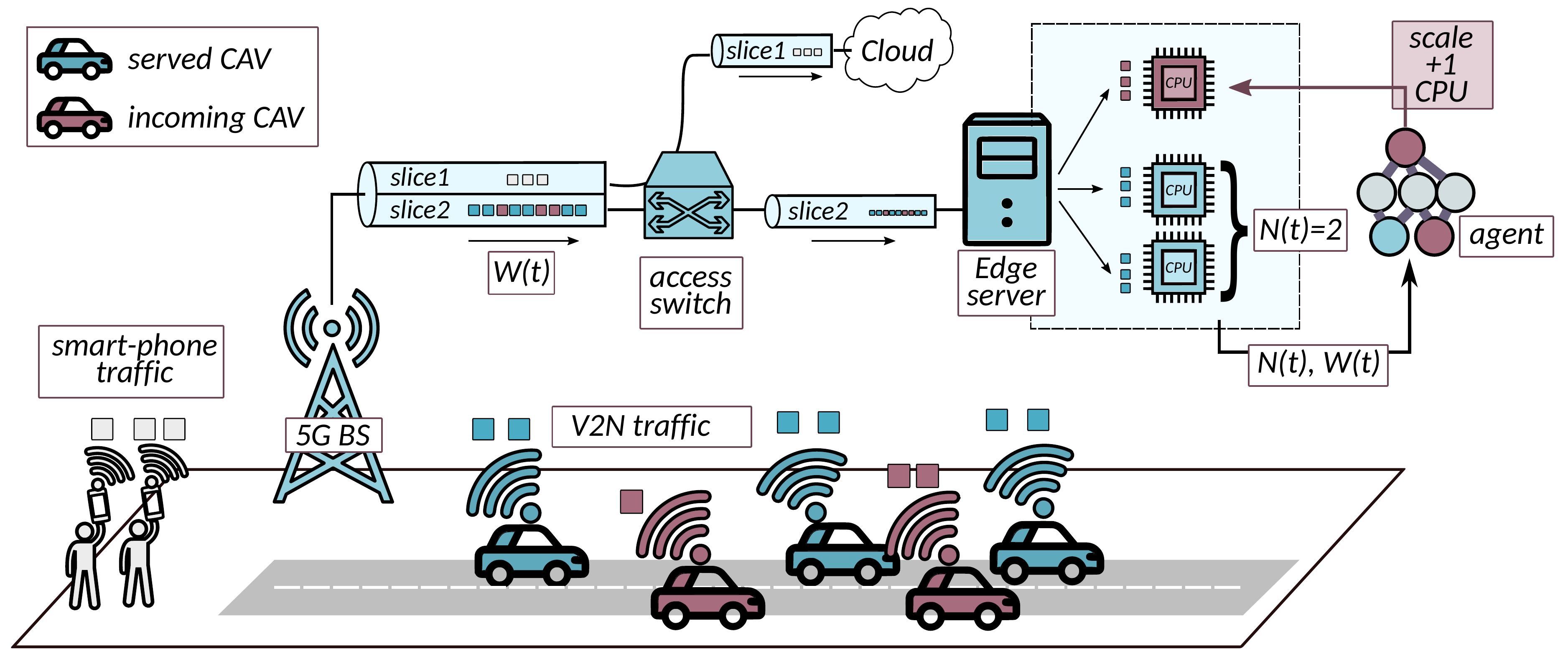}
    \caption{Considered V2N system.}
      \vspace{-1.5em}
    \label{fig:system}
\end{figure}

 For the V2N system, we consider a road segment in the coverage area of a 5G base-station (BS) as depicted in Fig.~\ref{fig:system}. The BS provides connectivity to smartphone users and CAVs along the road.
 We assume that every CAV uses V2N-based applications such as remote driving, hazard warning etc., and its traffic is processed in the edge of the network to satisfy latency requirements.  We assume smartphones and CAVs are connected to their respective slices i.e., \emph{slice~1} supporting smartphones' traffic processed in the cloud; and \emph{slice~2} supporting V2N traffic processed at the Edge server. 
 
 We focus on the workload 
 $W_t\in\mathbb{R}^+$ that V2N
 services introduce in \emph{slice 2}
 over time $t$. 
 If there are $V_t$~vehicles, each
 of them sending $P_v$~packets/sec,
 and each CPU~$c_i$ processes
 $P_{c_i}$~packets/sec on time
 (i.e., satisfying latency requirements);
 then the  workload is expressed as
 $W_t=P_v V_t/P_{c_i}$.
The goal is that the system in
 Fig.~\ref{fig:system}~distributes the
 overall V2N workload $W_t$ among the Edge CPUs
 to satisfy latency requirements, i.e.,
 the workload $W_t^{c_i}$ 
 dispatched to CPU $c_i$
 should satisfy $W^{c_i}_t\leq 1$.
 Thus, the system should efficiently
 (vertically) scale the number of CPUs
 $N_t\in\mathbb{N}^+$ by turning them on/off to process the V2N traffic on time. To that end, we propose using
 an ML agent (see~Fig.~\ref{fig:system})
 that learns the traffic patterns, and
 anticipates workload fluctuations, meeting delay requirements by scaling up/down the number of CPUs.



\section{Problem statement and Proposed Approach}
\label{sec:problem}


In this section we first
discuss the MDP associated to the V2N system described in \S\ref{sec:system}. Then describe how we
use DRL to solve the MDP and scale the V2N service.

\vspace{-1em}
\subsection{Markov Decision Process}
\label{subsec:mdp}

Typically, MDPs are characterized by 
a tuple {$(\mathcal{S}, \mathcal{A}, \mathbb{P}_{a_t}, R)$}
denoting the inherent state space $\mathcal{S}$,
action space $\mathcal{A}$,
transition probability $\mathbb{P}_{a_t}$, and
reward function $R$.

\textbf{State Space.} The state $s_t$ at time $t$ is specified
by the number of active CPUs at the Edge server $N_t$, and the
workload $W_t$ associated with the incoming V2N traffic, i.e., the state is defined as the tuple $s_t=(N_t,W_t)$.

\textbf{Action Space.}
To increase/decrease or maintain the number of CPUs in the Edge server,
we define an action as \mbox{$a_t\in\mathcal{A}=\{-N_{\max}, \ldots, N_{\max}\}$},
with $N_{\max}$ being the maximum number of CPUs in the Edge server.


\textbf{Transition Probability.}
Given the current action $a_t$
and state $s_t=(N_t,W_t)$,
the transition
probability to the next state
$s_{t+1}=(N_{t+1},W_{t+1})=(N_t+a_t,W_{t+1})$
is determined
by $\mathbb{P}_{a_t}(s_{t+1}|\ s_t)=\mathbb{P} (s_{t+1}|\ a_t,s_t)$.

\textbf{Reward.}
The reward function $R$ depends on the workload $W_t^{c_i} = \min\big(1,\ x_i\cdot W_t + B_{t-1}^{c_i} \big)$ and backlog $B_t^{c_i} = \max \big(0,\ W_t^{c_i}-1\big)$~\cite{de20215growth}.
The workload $W_t^{c_i}$ of a CPU $c_i$ is
only a portion $x_i\in[0,1]$ of the
total workload $W_t$ plus its
prior backlog $B_{t-1}^{c_i}$, i.e.,
the workload that CPU $c_i$ could not
process. As mentioned in
\S\ref{sec:system}, the CPU workload
should remain below one to
satisfy V2N latency requirements,
thus clipping at one.
Based on the workload and
backlog definitions, we define
the reward as:
\begin{equation}
    R(s_{t+1}|\ s_t,a_t) = \min\{W_{t+1}^{c_i}\}_{i=0}^{N_t+a_t} - \beta\cdot\max\big\{B_{t+1}^{c_i} \big\}_{i=0}^{N_t+a_t}\label{eq:reward}
\end{equation}
which aims to maximize the CPU utilization
by encouraging the least loaded CPU to carry more
workload (first term), while penalizing the maximum backlog accumulated by a CPU (second term). The backlog penalty is weighted by a term $\beta\in\mathbb{R}^+$ to control its impact on the reward function. 


With the definition of the state and
action space, transition probabilities and reward function, we formulate the
MDP:
\vspace{-0.5em}
\begin{problem}[\bfseries V2N scaling MDP]
    \label{problem}
    Given the {$(\mathcal{S}, \mathcal{A}, \mathbb{P}_{a_t}, R)$}
    tuple, find a policy $\pi$ that maximizes:
    \begin{equation}
         \mathbb{E}_{\substack{\hspace{-1em}a_t\sim\pi,\\\hspace{-.7em}s_{t+1}\sim\mathbb{P}_{a_t}(s_t)} } \Big[ \sum_t \gamma^t \big( \min\{W_{t+1}^{c_i}\}_{i=0}^{N_t+a_t}- \beta\cdot\max\big\{B_{t+1}^{c_i} \big\}_{i=0}^{N_t+a_t} \big)  \Big] 
        \label{eq:expected-reward}
    \end{equation}
    with $\gamma\in[0,1]$ being the discount factor.
\end{problem}
In other words, we aim to find an optimal policy
$\pi$ to maximize the expected discounted reward. We resort to a model-free RL 
approach to find an optimal policy
$\pi$ without making assumptions about the transition probabilities. 


\vspace{-1em}

\subsection{V2N scaling with DDPG-DOD}
\label{subsec:a2cddpg}
RL finds an optimal scaling policy $\pi$ for \textit{Problem~\ref{problem}}
using an estimation of the expected discounted
reward~\eqref{eq:expected-reward}. Such estimation is
known as the expected gain
$\mathbb{E}[G_t]=\mathbb{E}\left[\sum_t\gamma^{t} R(s_{t+1}| s_t,a_t)\right]$,
and an optimal policy $\pi$ will take the
adequate scaling actions $a_t$ to maximize
$\mathbb{E}[G_t]$. In this section we advocate for
the following
DRL agent to estimate $\mathbb{E}[G_t]$ and
look for optimal scaling policies $\pi$
for \textit{Problem~\ref{problem}}.

\textbf{DDPG-DOD.} 
DDPG \cite{DBLP:journals/corr/LillicrapHPHETS15} is an RL algorithm that draws from deterministic policy gradient and DQN for learning in continuous action space.
Similar to A2C\cite{10.5555/3045390.3045594}, DDPG 
is based on the actor-critic architecture, where the critic approximates the expected gain $\mathbb{E}[G_t]$ with the action-value function $Q(s_t,\hat{a_t})=\mathbb{E}[G_t|s_t, \hat{a_t}]$, $\hat{a_t}\in\mathbb{R}$.
However, unlike the A2C actor that estimates the probability distribution of actions $\pi(a_t|s_t)$, the DDPG actor learns a deterministic policy $\pi(s_t)$ which generates a real-valued action $\hat{a_t}$.
DDPG is not directly applicable to scaling problems with discrete actions. To map the real-valued action $\hat{a_t}$ to the number
of CPUs to scale up/down, we propose a transformation called
\textit{Deterministic Ordered Discretization} (DOD).
Given the real-valued output of DDPG $\hat{a_t}\in[l,u]$, then DOD is a transformation
\begin{equation}
    g(\hat{a_t}) = \operatorname*{argmin}_{a\in\mathcal{A}} \left\| a \ - \left(  \hat{a_t} \frac{2N_{\max}}{u-l} - N_{\max}\frac{u+l}{u-l} \right) \right\|
    \label{eq:dod}
\end{equation}
that ($i$) applies
a positive affine transformation that maps the real-valued
action $\hat{a_t}$ from the range $[l,u]$ to 
$[-N_{\max},N_{\max}]$; and ($ii$) finds the
nearest discrete action $a\in\mathcal{A}$. Using DOD with DDPG presents two
advantages:
\begin{enumerate}
    \item \emph{DOD mitigates the explosion of the
    action space}. Regardless of the number of available
    CPUs, DOD always takes a single real-valued action $\hat{a_t}$ given by DDPG and maps it to the discrete action $a_t \in \{ -N_{\max},\ldots, N_{\max} \}$.
    Scaling DRL solutions in the literature
    \cite{gari2021reinforcement}
    use as many neurons for the output layer
    as number of CPUs,
    which makes the output dimension grows as $\mathcal{O}(N_{\max})$.
    
    \item \emph{DOD exploits the internal ordering of the problem}.
    If the certain action $a_t$ has a higher chance to be selected given the state $s_t$, its proximate actions (e.g. $a_t\pm1$) also get higher chances. 
     Learning can become more efficient by leveraging such relations between actions, while typical discrete action RL algorithms (e.g. DQN, A2C) take each action as an independent option, thus the structure of the action space is ignored. The authors in
    \cite{dulacarnold2016deep,tang2020discretizing} have proposed different approaches to support a similar idea.
\end{enumerate}

\noindent  Note that we can still update DDPG-DOD agent via policy gradient, as DOD can be seen as a part of the environment (i.e. a step prior to the reward function calculation), which does not play a role in the gradient update.

\section{Performance Evaluation}
\label{sec:results}
\subsection{Workload Generation}
\label{dataset}
We consider a real-world dataset with a traffic trace from Corso Orbassano road in Turin, spanning from January 2020 to October 2020. The trace contains the number of cars that pass via certain measuring points every 5 minutes. We split the complete trace in 80:20 ratio for training and testing purposes respectively. 
Following the assumptions in~\cite{de20215growth},
 $V_t=8$~vehicles using a video-related V2N service (e.g., remote driving) generate a workload of $W_t=1$.
In other words, a single CPU processes on
time the traffic sent by 8~vehicles, i.e., $P_c=8P_v$.
We assume that the total workload $W_t$ is distributed across the different CPUs according to the Dirichlet distribution. That is, the load $x_i$ for each CPU $c_i$ satisfies $\sum_i x_i=1$, 
while \mbox{$\mathbb{P}(x_1,\ldots,x_{N_{\max}})\sim \prod_{i}x_i^{\alpha_i-1}$}. With
$\alpha_i=1000$ in our evaluation environment, the
workload $W_t$ is almost evenly distributed among the
CPUs. The weight of the backlog penalty in the reward function is set to $\beta=1.0$ \cite{de20215growth} and the discount factor is set to $\gamma=0.99$.
\subsection{State of the
art solutions}
\label{ScalingApproaches}
We compare our scaling approach to other solutions, as presented in \cite{de20215growth} and \cite{beks21}, namely:
\setlist{nolistsep}
\begin{itemize}[noitemsep,leftmargin=*]
    \item  a \textbf{Proportional Integral} (PI) controller \cite{Ang05pidcontrol} 
    that aims to keep the most loaded CPU below a threshold of $\rho=0.6$; 
    \item a \textbf{Long Short-Term Memory} (LSTM) predictor \cite{hochreiter1997lstm}
    with 2 layers with 4 cells each, where we use a look back of 3 slots (i.e., a prediction is based on the 3 previous values); 
    \item a \textbf{Q-Learning} (RL) algorithm
    \cite{Watkins1992qlearning} 
    with the same state space and reward function as the proposed one, but only three actions: \mbox{$a_t\in\mathcal{A}=\{-1,0,1\}$}
   \item an \textbf{A2C} based scaling approach \cite{10.5555/3045390.3045594}. To have a fair comparison, we apply similar experimental setup as that of DDPG-DOD, as described in the following subsections.
\end{itemize}

\begin{figure}[b]
     \centering
     \begin{subfigure}[b]{0.45\columnwidth}
         \centering
         \includegraphics[width=\textwidth]{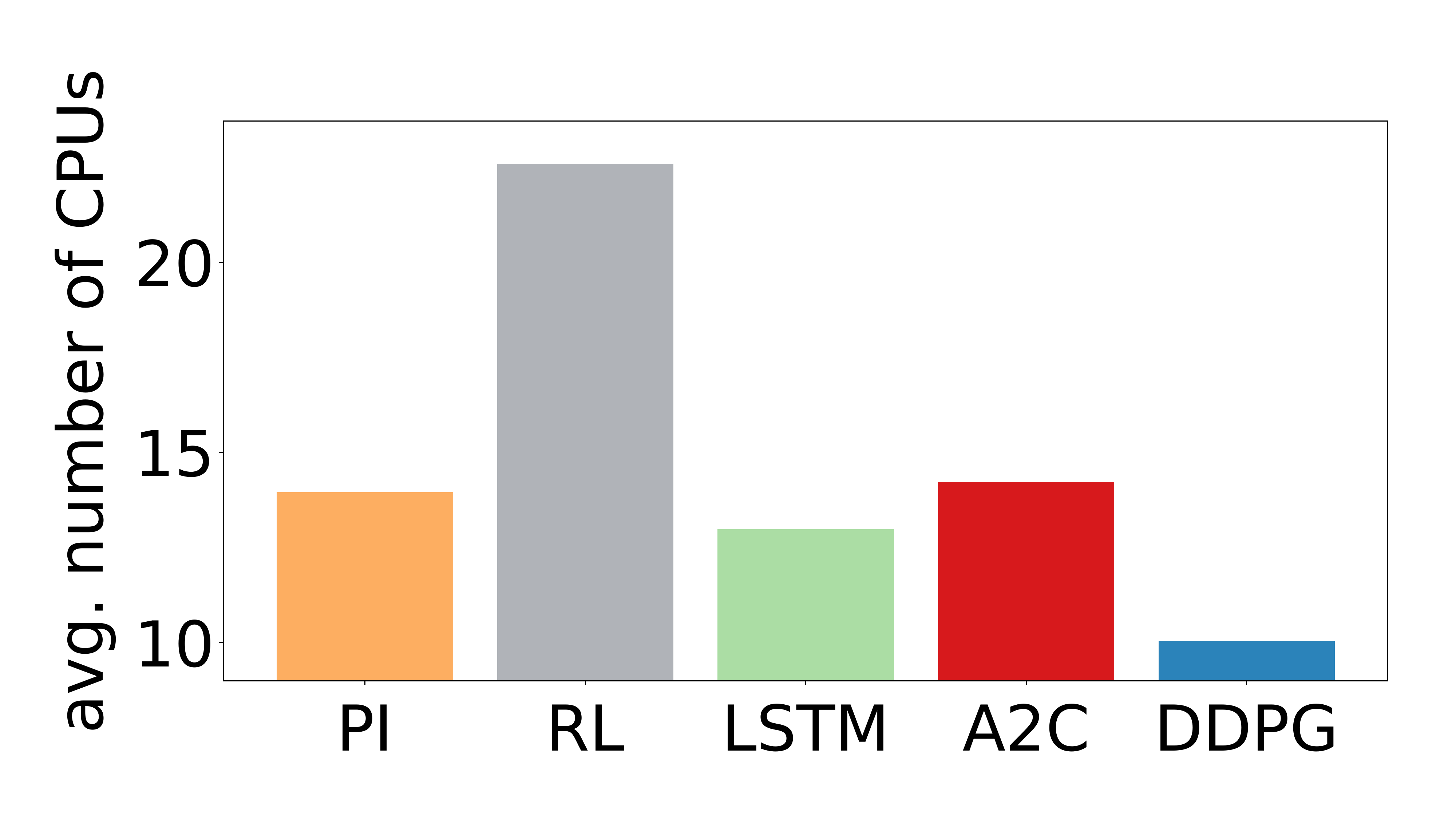}
         \caption{Number of CPUs}
     \end{subfigure}
     \begin{subfigure}[b]{0.45\columnwidth}
         \centering
         \includegraphics[width=\textwidth]{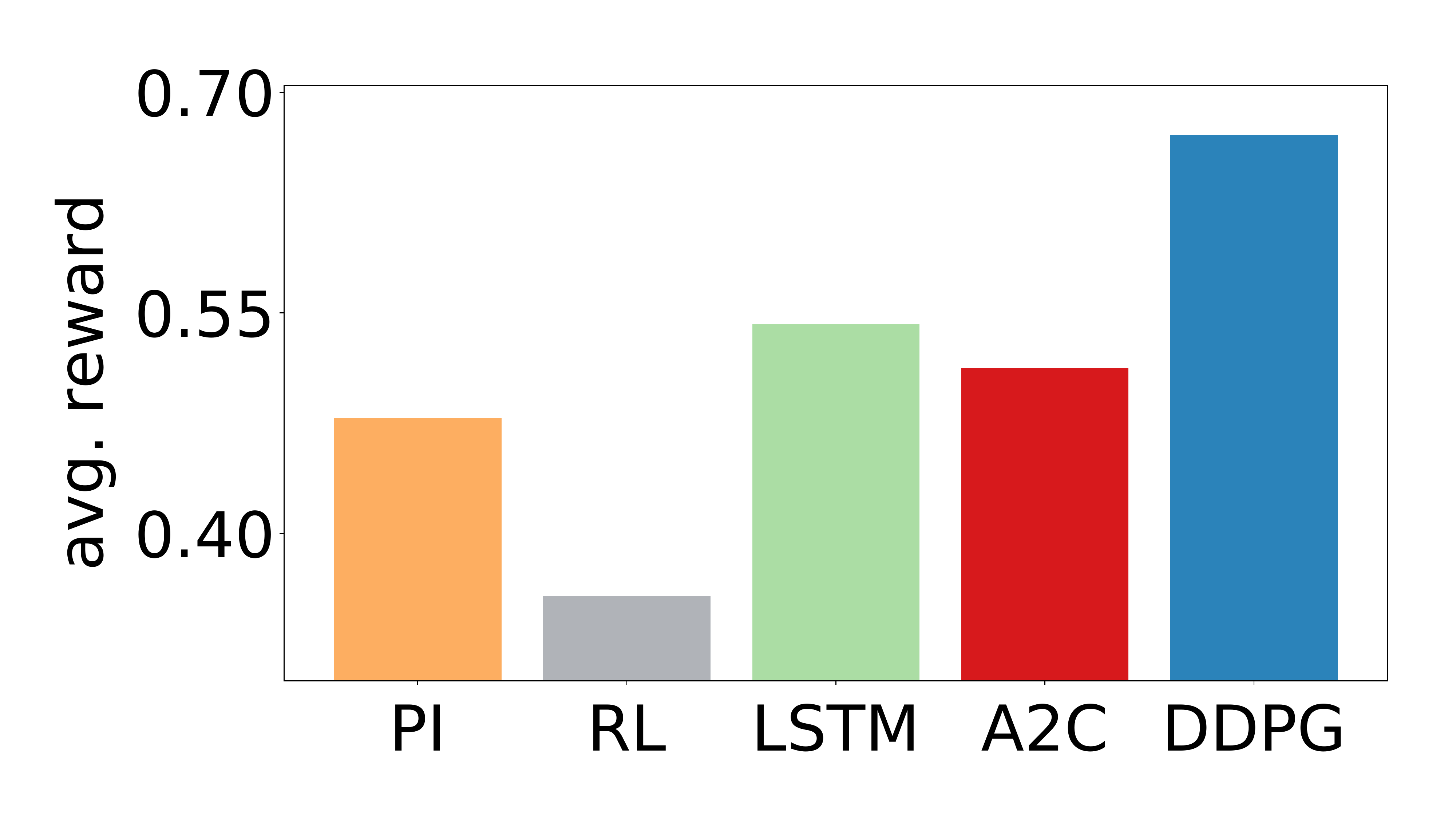}
         \caption{Reward}
     \end{subfigure}
        \caption{Average performance metrics}
        \label{fig:baseline}
\end{figure}
\vspace{-1em}
\subsection{A2C and DDPG-DOD setup}
\label{EvalEnv}

The actor and critic networks of DDPG-DOD and A2C
are multi-layer perceptrons (MLPs) with 3 hidden layers of 128 neurons.
Learning rate is set to $lr=3e-3$.
The code implemented by the authors in \cite{de20215growth} is used for the environment. We implement A2C and DDPG-DOD with PyTorch 1.10.0. The tests are carried out on a server with an Intel Core i7-10700K CPU and 32 GB of RAM. 
\subsection{Metrics and evaluation scenarios}
\label{metr}
Our goal is to maximize the long
term reward of \textit{Problem~\ref{problem}},
which minimizes the operational cost via a proper scaling of
computational resources over time.
Here we use the \textit{average number of active CPUs} and the \textit{average reward} as metrics. 
We set up two different evaluation
scenarios to assess the performance
of the different solutions:
($i$) a \emph{Performance}
scenario where we test every
solution over two
days in Corso Orbassano using an
action space
$\mathcal{A}=\{-5,\ldots,5\}$, $\mathcal{|A|}=11$; and
($ii$) a \emph{Scalability} scenario
where we study the impact of increasing the action space up to
$\mathcal{A}=\{-15,\ldots,15\}$, $\mathcal{|A|}=31$,
and $\mathcal{A}=\{-25,\ldots,25\}$, $\mathcal{|A|}=51$.



\subsection{Results}
\label{comp}

\begin{figure}
     \centering
     \begin{subfigure}[b]{0.45\textwidth}
         \centering
         \includegraphics[width=\textwidth]{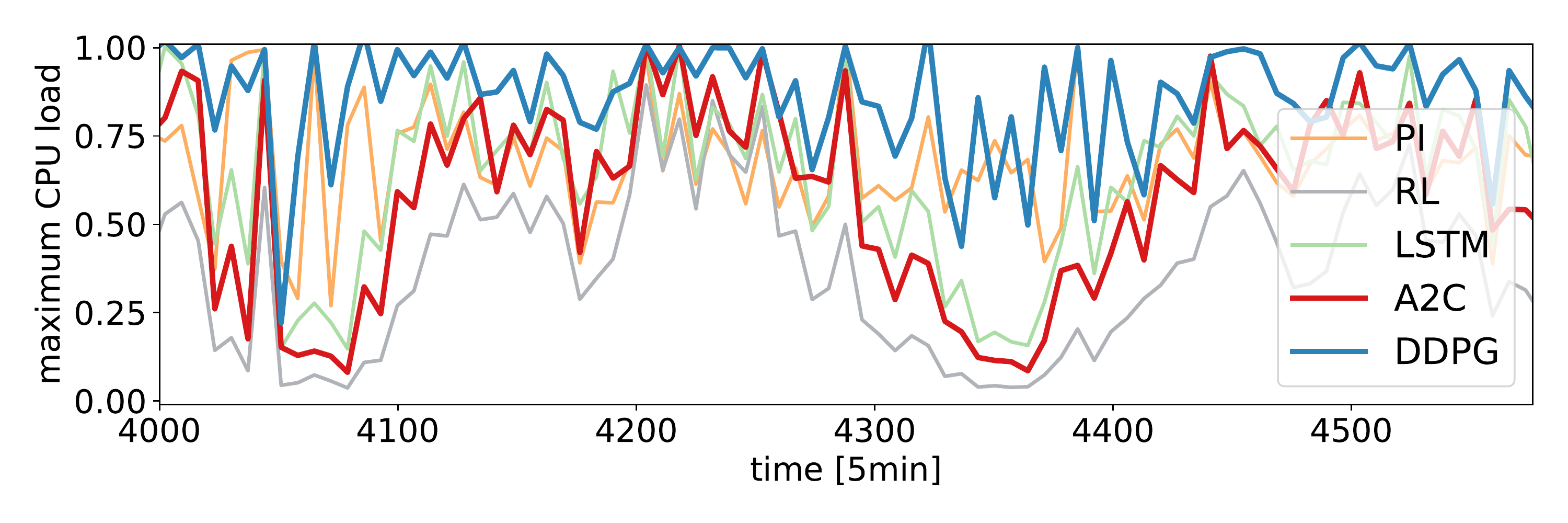}
         \caption{Maximum load of CPUs}
          \label{performaceA}
     \end{subfigure}
     \centering
     \begin{subfigure}[b]{0.45\textwidth}
         \centering
         \includegraphics[width=\textwidth]{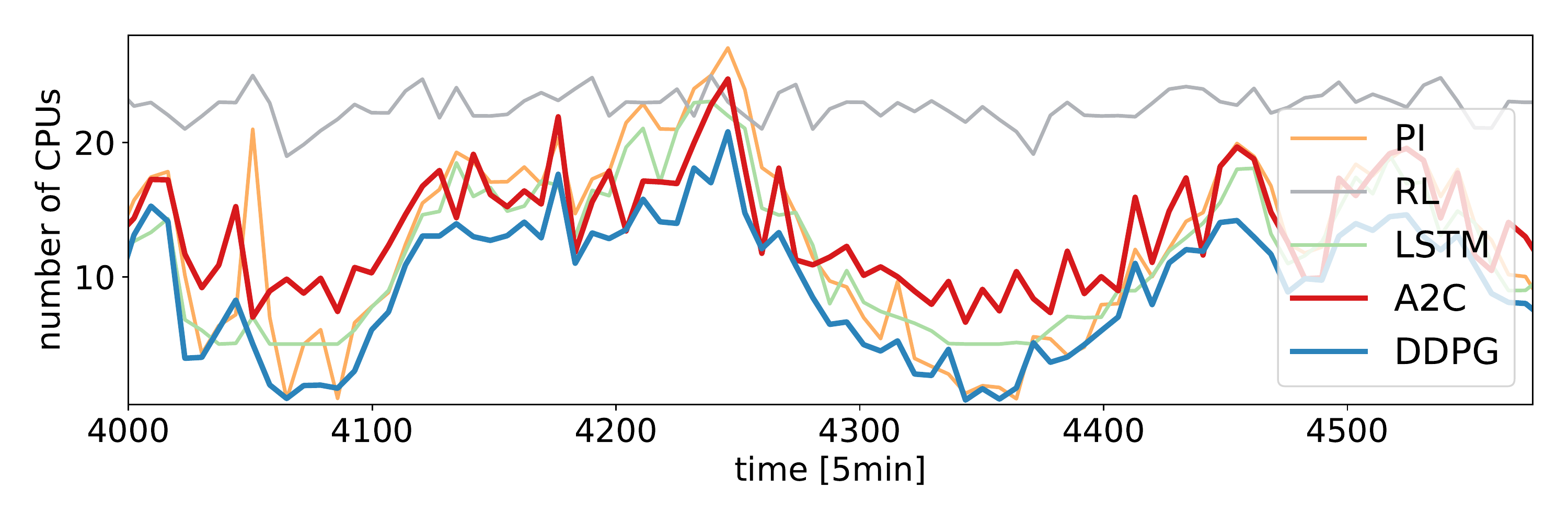}
         \caption{Number of CPUs}
          \label{performaceB}
     \end{subfigure}
     \begin{subfigure}[b]{0.45\textwidth}
         \centering
         \includegraphics[width=\textwidth]{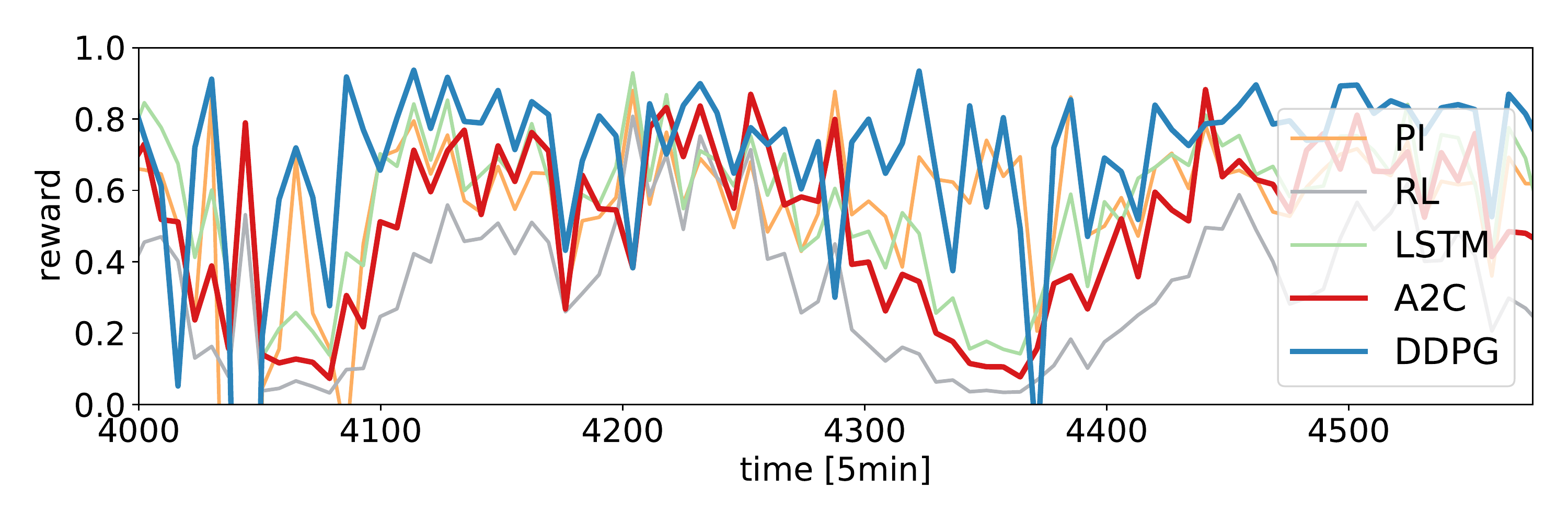}
         \caption{Reward}
          \label{performaceC}
     \end{subfigure}
    \caption{Trace of performance over time}
    \vspace{-1.8em}
    \label{fig:trace}
\end{figure}

\textbf{Performance.} 
Fig.~\ref{fig:baseline} plots the average number of CPUs and reward that each solution obtains
over the testing trace. 
The Q-learning agent, denoted as RL, maintains on average the largest number of active CPUs, that leads to a reduction in the average reward (first term in formula~\eqref{eq:reward}). The A2C agent performs similar to the PI controller, exhibiting slightly higher average reward (6\% increase) for marginally higher number of CPUs (2\%). LSTM and DDPG-DOD (denoted hereafter as DDPG for the sake of simplicity) tend to attain larger rewards while keeping the average number of the CPUs low, which essentially reduces the overall cost for the edge resources used. The increased average reward also indicates a reduced backlog for the particular approaches -- as indicated by the second term in formula~\eqref{eq:reward}. However, DDPG outperforms LSTM, decreasing by approximately  23\% the average number of CPUs, while increasing by 24\%  the average reward.



Fig.~\ref{fig:trace} illustrates 
 how all approaches perform  over a period of two days regarding the maximum CPU load $\max_i\{W_t^{c_i}\}$, number of active CPUs $N_t$ and reward $R(s_{t+1}|\ s_t,a_t)$. The RL agent is the most conservative agent; it maintains a workload of $W_t^{c_i}<1$ for all CPUs (Fig.~\ref{performaceA}) with allocating the largest number of active processors (Fig.~\ref{performaceB}), which leads to a reduction in the  reward (Fig.~\ref{performaceC}).
As discussed in the previous paragraph the A2C agent performs similar to the PI controller.
However in Fig.~\ref{performaceA} we note that A2C avoids overloading the resources compared to PI, while it increases the maximum load  compared to more conservative approaches like RL.
 We further observe in Fig. \ref{fig:trace} that DDPG and PI are more aggressive solutions; 
 they activate a sufficient number of CPUs while frequently a CPU gets overloaded, even only for short time period, especially in the case of DDPG (Fig.~\ref{performaceA}).
 However DDPG outperforms PI both in terms of reward as well as cost (average number of active CPUs) as presented also in Fig. \ref{fig:baseline}. Moreover the number of CPUs over time for DDPG (Fig.~\ref{performaceB}) evolves smoother compared to the PI case.
We argue that the decision boundary of DDPG is smoother than that of the discrete-action approaches as the discretization method takes into account the ordering property of the scaling action space. 
LSTM and DDPG are the most efficient approaches in terms of the average values of the selected metrics, as discussed in the previous paragraph (Fig. \ref{fig:baseline}).
Note that in Fig.~\ref{performaceB} DDPG reduces further the number of CPUs compared to LSTM, which leads to the high variation in the maximum load depicted in Fig.~\ref{performaceA}.
However, Fig.~\ref{performaceC} suggests that DDPG keeps the periods with fully loaded CPUs 
short, thus, the backlog incurred has low impact on the reward.



\begin{figure}[t]
     \centering
     \begin{subfigure}[b]{0.45\columnwidth}
         \centering
         \includegraphics[width=\textwidth]{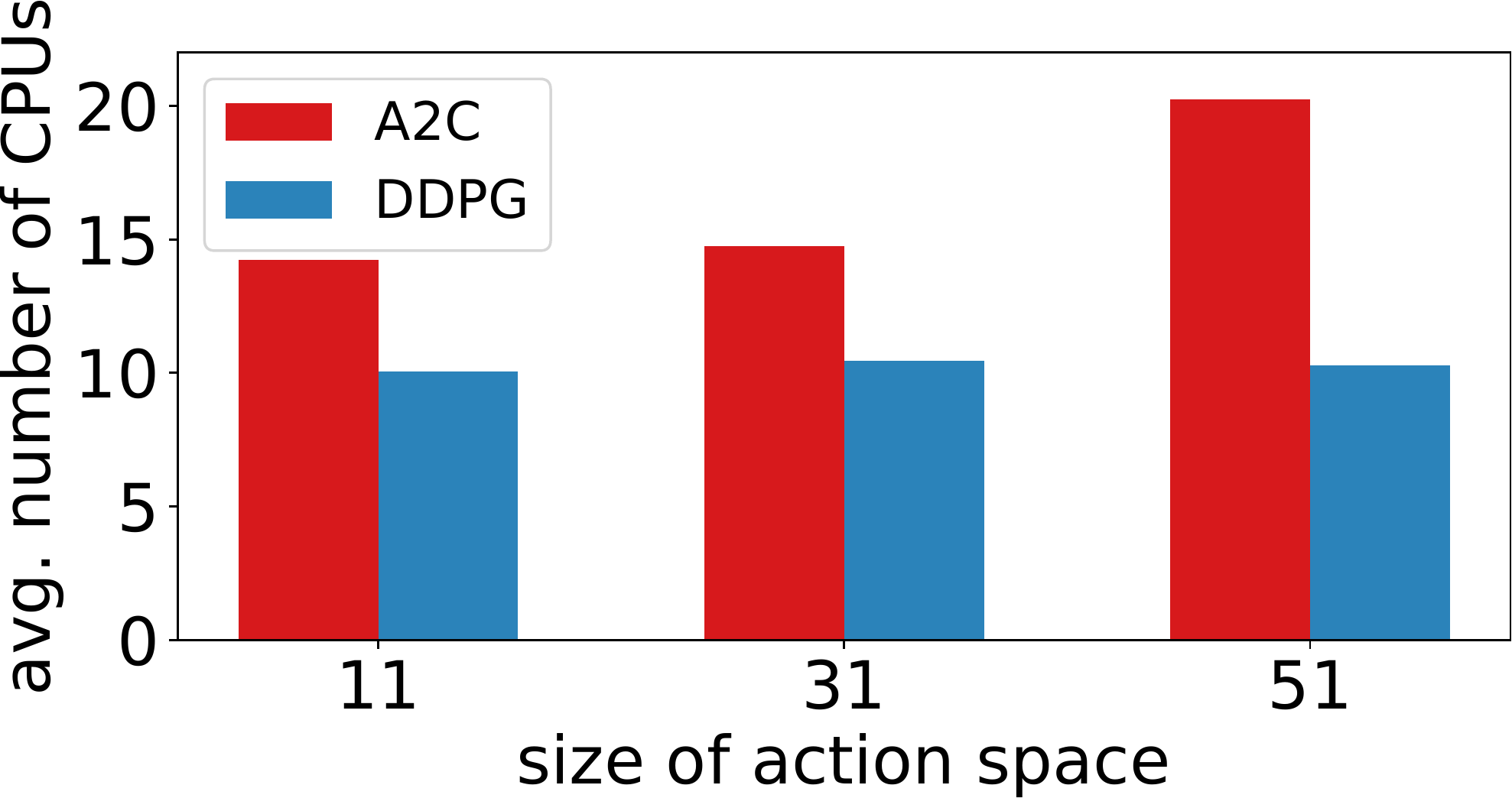}
         \caption{Number of CPUs}
     \end{subfigure}
     \hspace{.2em}
     \begin{subfigure}[b]{0.45\columnwidth}
         \centering
         \includegraphics[width=\textwidth]{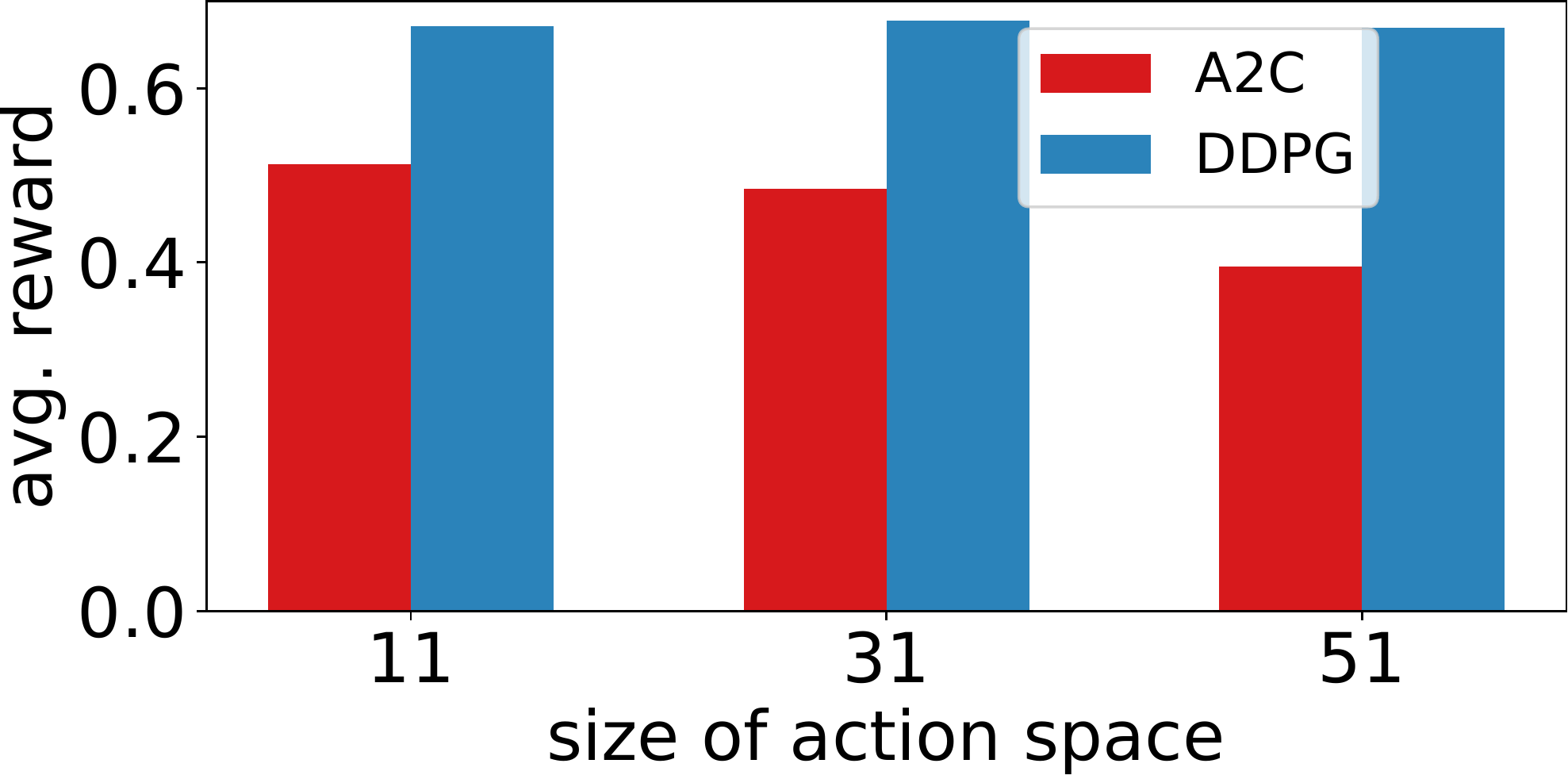}
         \caption{Reward}
     \end{subfigure}
        \caption{Scalability over various sizes of action space}
          \vspace{-1em}
        \label{fig:scalability}
\end{figure}

\textbf{
Scalability.} We look into the scalability of the DRL agents regarding the size of the action space. Fig. \ref{fig:scalability} compares the average reward and the average number of CPUs that are allocated by A2C and DDPG with various dimensions of the action space. 
The results show that DDPG performs robustly as the size of the action space increases, while for A2C it gets more difficult to converge.
DDPG converges in spite of the larger
action spaces by exploiting the underlying order between discrete actions.
In contrast, A2C takes each action as an independent option. As the structure of the action space is not utilized, thus the learning process is less efficient.
 
 We have also studied how A2C works with the ordinal architecture proposed in~\cite{tang2020discretizing}.
In this case, A2C failed to converge with an action space of size 31 or 51, due to the highly interleaved logits in the forward/backward passes~\cite{tang2020discretizing} that lead to high memory usage. The converged configuration, for an action space of size 11, performs on average worse than the original A2C with a reward of 0.46 and 14.34 CPUs.


\section{Conclusion}
\label{sec:conclusions}
In this paper we propose to vertically
scale V2N services using DDPG-DOD, a DDPG agent equipped with
a non-parametric discretization method that is designed to capture the structure
of the scaling decisions and learn discrete actions in a continuous fashion -- thus avoiding the action-space explosion.
Employing a real-world vehicular trace dataset, we show  that DDPG-DOD outperforms state of the art solutions in terms of (i) operational cost as it minimizes the number of active CPUs, (ii) performance;  
increasing the long-term reward is an indicator of reduced backlog and thus processing delay, and (iii) flexibility in scaling resources as DDPG-DOD performs robustly independently of the size of the action space.

In future work, we plan to investigate the applicability of the proposed approach for V2N service scaling in a Multi-PoP environment, covering a metropolitan area. In such environment with more degrees of freedom in offloading computations, placement decisions should be also taken into consideration. 


\section*{Acknowledgements}
This work has been partially funded by
the Spanish Ministry of Economic Affairs and Digital
Transformation and the European Commission through 
the 6G-EDGEDT, 6G-DATADRIVEN and DESIRE6G (grant no. 101095890) projects.

\balance
\bibliographystyle{IEEEtran}
\bibliography{reference.bib}

\end{document}